\documentclass[runningheads]{llncs}

\usepackage{eccv}

\usepackage{eccvabbrv}

\usepackage{graphicx}
\usepackage{booktabs}
\usepackage{wrapfig}
\usepackage[accsupp]{axessibility}  

\usepackage{hyperref}

\usepackage{orcidlink}

\usepackage{url}
\usepackage{soul}
\usepackage{multicol}
\usepackage{multirow}
\usepackage{array}
\usepackage{wrapfig}
\usepackage{tabularx}
\usepackage{graphicx}
\usepackage{colortbl}
\usepackage{amsmath, mathtools}
\usepackage[ruled,vlined,linesnumbered]{algorithm2e}
\usepackage{enumitem}
\usepackage{booktabs} 
\usepackage{capt-of}
\usepackage{fontawesome}
\usepackage{bm}
\usepackage{dsfont}
\usepackage{circledsteps}
\usepackage{algorithm2e}
\usepackage{algpseudocode}

\definecolor{mygreen}{RGB}{0,140,0}
\definecolor{myblue}{HTML}{FDF5E0}

\newcommand{\ours}{DSeq-JEPA}
\newcommand{\cours}{DSeq-C-JEPA}
\definecolor{lightblue}{rgb}{0.8,0.9,0.95}

\begin{document}

\title{DSeq-JEPA: Discriminative Sequential Joint-Embedding Predictive Architecture} 

\titlerunning{DSeq-JEPA}


\author{Xiangteng He\textsuperscript{*}\inst{1,2}
\and
Shunsuke Sakai\textsuperscript{*}\inst{3} \and
Shivam Chandhok\inst{1,2} \and
Sara Beery\inst{4} \and \\
Kun Yuan\inst{5,6} \and 
Nicolas Padoy\inst{5,6} \and
Tatsuhito Hasegawa\inst{3} \and
Leonid Sigal\inst{1,2,7}}
%

\authorrunning{He et al.}


\institute{$^{1}$University of British Columbia, Canada, $^{2}$Vector Institute for AI, Canada, 
$^{3}$University of Fukui, Japan $^{4}$Massachusetts Institute of Technology, USA, $^{5}$University of Strasbourg, France, 
$^{6}$IHU Strasbourg, France, 
$^{7}$Canada CIFAR AI Chair, Canada
}

\maketitle
\vspace{-0.5em}
\begin{center}
{\small \textbf{Project Page:} \url{https://dseqjepa-project.com}}
\end{center}
\vspace{-0.5em}

\begingroup
\renewcommand\thefootnote{}
\footnotetext{* Equal contribution.}
\endgroup

\begin{abstract}
Recent advances in self-supervised visual representation learning have demonstrated the effectiveness of predictive latent-space objectives for learning transferable features.
In particular, Image-based Joint-Embedding Predictive Architecture (I-JEPA) learns representations by predicting latent embeddings of masked target regions from visible context. However, it predicts target regions in parallel and all at once, lacking ability to order predictions meaningfully. 
Inspired by human visual perception, which attends selectively and progressively from primary to secondary cues, we propose {\sc \ours}, a {\sc D}iscriminative {\sc Seq}uential {\sc J}oint-{\sc E}mbedding {\sc P}redictive {\sc A}rchitecture that bridges latent predictive and autoregressive self-supervised learning. Specifically, \ours~integrates a discriminatively ordered sequential process with JEPA-style learning objective. 
This is achieved by (i) identifying primary discriminative regions using an attention-derived saliency map that serves as a proxy for visual importance, and (ii) predicting subsequent regions in discriminative order, inducing a curriculum-like semantic progression from primary to secondary cues in pre-training. Extensive experiments across tasks -- image classification (ImageNet), fine-grained visual categorization (iNaturalist21, CUB, Stanford Cars), detection/segmentation (MS-COCO, ADE20K), and low-level reasoning (CLEVR) -- show that \ours~consistently learns more discriminative and generalizable representations compared to I-JEPA variants.
\end{abstract}

\section{Introduction}
\label{sec:introduction}

Self-supervised learning (SSL) has become a powerful paradigm for learning robust and transferable visual representations from unlabeled data. 
Early advances in contrastive learning~\cite{simclr,moco,mocov2} and masked image modeling~\cite{mae,beit,simmim} showed that representations learned without manual annotation can rival supervised pretraining on many downstream tasks. 
More recently, Joint-Embedding Predictive Architectures (JEPAs)~\cite{jepa} have shifted the objective from contrasting instances or pixel reconstruction to latent-space prediction, offering a scalable and information-preserving formulation.
Among these, I-JEPA~\cite{ijepa} has emerged
as a compelling framework for learning high-level visual representations through latent prediction of masked target regions.
Despite its effectiveness, I-JEPA treats target regions largely uniformly and independently during pre-training. 

\begin{figure}[t]
{
    \centering
    \includegraphics[width=\textwidth]{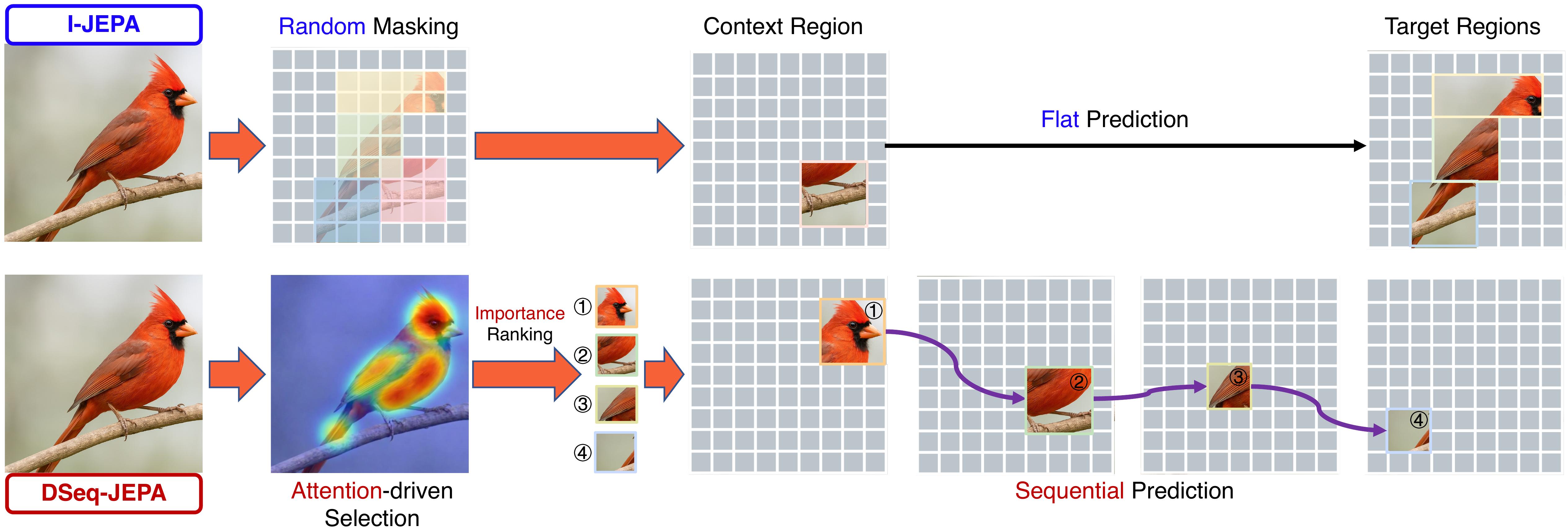}
    \caption{
    \textbf{Overview of the key differences between I-JEPA and \ours.} I-JEPA predicts embeddings of randomly sampled target regions in a parallel and independent manner. In contrast, \ours~ranks regions by attention and predicts region embeddings sequentially from high- to low-attention regions. Patches shown are only for illustration; the model predicts region embeddings rather than raw pixels.
    }
    \label{fig:motivation}
}
\end{figure}

In practice, visual information is rarely uniform: some regions carry primary semantic cues (\eg, object-defining parts), while others provide secondary or contextual evidence. Moreover, visual perception is often \emph{selective} and \emph{sequential}, attending first to salient cues and then refining interpretation using additional context.
This motivates a key research question: 


\begin{quote}\small\itshape
Can predictive self-supervised learning benefit from an explicit notion of \ul{\textbf{where}} and \ul{\textbf{in what order}} to make predictions?
\end{quote}


In this work, we answer this question affirmatively and propose Discriminative Sequential Joint-Embedding Predictive Architecture (\ours), a new self-supervised pretraining framework that introduces a discriminative sequential inductive bias that is (loosely) motivated by human perception. 
This paradigm bridges two previously distinct lines of self-supervised learning:
\emph{predictive modeling} (JEPA-style) and \emph{auto-regressive next region prediction} (iGPT-like~\cite{igpt}).
Our key idea is to impose structure on target predictions by prioritizing semantically informative regions and predicting them in a progressive order.
As illustrated in \Cref{fig:motivation}, \ours~first derives an attention-based saliency map to estimate visual importance and identify primary discriminative regions (\emph{where to attend}), instead of I-JEPA's uniform sampling. It then performs sequential latent predictions over these regions from the most discriminative to the least (\emph{in what order}), rather than I-JEPA’s flat, independent predictions. Ultimately, \ours~yields a curriculum-like semantic progression  from primary to secondary cue during pre-training. 

The proposed design is motivated by two complementary observations. First, \emph{where} the model predicts matters: focusing on discriminative regions can concentrate learning on informative content instead of treating all targets equally. Second, \emph{in what order} the model predicts matters: sequential prediction encourages progressive representation building, where earlier high-value cues support subsequent predictions. Importantly, these components are synergistic: discriminative region selection provides an informative ordering prior, and sequential latent prediction converts this prior into a structured self-supervised learning signal. From this perspective, \ours~can be viewed as breaking the permutation symmetry over target regions implicit in independent region prediction and replacing it with a semantically meaningful prediction order.

We validate the effectiveness of \ours~through extensive experiments across diverse downstream tasks and evaluation settings. Compared with I-JEPA variants, \ours~consistently improves transfer performance on image classification (ImageNet), fine-grained visual categorization (iNaturalist21, CUB-200-2011, Stanford Cars), dense prediction tasks including detection and segmentation (MS-COCO, ADE20K), and low-level reasoning benchmarks (CLEVR). Beyond quantitative gains, qualitative analyses show that \ours~attends to and predicts more discriminative regions, while ablations demonstrate that both discriminative region selection and sequential latent prediction are necessary and that their combination yields the strongest improvements.

\vspace{0.1in}
\noindent
{\bf Contributions:} Our contributions can be summarized as follows:
\begin{enumerate}[leftmargin=*,topsep=2pt,itemsep=1pt]
    \item[-] We introduce \textbf{\ours}, a JEPA-based self-supervised pre-training framework that augments latent predictive learning with \textbf{discriminative region prioritization} and \textbf{sequential next-region embedding prediction}.
    \item[-] We propose an attention-derived saliency-based mechanism to identify informative regions and construct a discriminative \textbf{prediction order}, enabling a curriculum-like progression from primary to secondary cues.
    \item[-] Through extensive experiments and ablations, we show that \ours~consistently improves over strong baselines across classification, fine-grained recognition, dense prediction, and low-level reasoning benchmarks.
\end{enumerate}

\noindent
By explicitly modeling both \textbf{where} and \textbf{in what order} to predict, \ours~advances joint-embedding predictive learning toward a more selective and progressive form of self-supervised visual pre-training.

\section{Related Work}
\label{sec:related}


Self-supervised learning (SSL) has become a central paradigm for learning transferable visual representations from unlabeled data. 
Existing SSL methods can be broadly grouped into three families: 

\paragraph{\textbf{Joint-Embedding Architectures}} are designed to bring the embeddings closer together for compatible inputs, and push the embeddings further apart for incompatible ones. This includes: (i) contrastive learning approaches, \textit{e.g.}, SimCLR \cite{simclr}, MoCo \cite{moco,mocov2,mocov3}, and SwAV \cite{swav}, which form the foundation of this paradigm, optimizing instance-level discrimination via large batches or memory banks; and (ii) non-contrastive approaches, \textit{e.g.}, BYOL \cite{byol}, SimSiam \cite{simsiam}, which further show that strong representations can emerge without explicit negatives via asymmetric prediction and momentum-based designs that avoid collapse. 

\paragraph{\textbf{Generative Architectures}} aim to reconstruct a target signal $\mathbf{y}$ (\emph{e.g.}, pixels, or patches) from a related input signal $\mathbf{x}$ by employing a decoder network. Masked image modeling serves as a prominent technique in this category, including MAE \cite{mae}, BEiT \cite{beit}, SimMIM \cite{simmim}, and iGPT \cite{igpt}, where models learn to fill in masked regions of an image. 
Although highly effective for representation learning, their objectives are driven by content reconstruction, which can allocate substantial modeling capacity to low-level appearance recovery rather than directly encouraging high-level semantic abstraction.
Notably, iGPT~\cite{igpt} adopts an autoregressive formulation that predicts image tokens sequentially.
Relatedly, RandSAC~\cite{randsac} explores self-supervision through random segments with autoregressive coding, suggesting that sequential prediction can serve as a useful inductive bias for visual representation learning.
However, these approaches do not explicitly model which regions are more informative and reconstruct pixels.

\paragraph{\textbf{Joint-Embedding Predictive Architectures}} formulate SSL as prediction in latent embedding space, \emph{i.e.}, predicting the embedding of a target signal from a compatible context signal through a predictor, rather than reconstructing pixels. By operating in the latent space, JEPAs encourage abstraction and provide a scalable predictive learning framework. I-JEPA~\cite{ijepa}, the image-based instantiation, predicts latent embeddings of masked target regions from visible context using a lightweight predictor, avoiding both pixel-level reconstruction and contrastive pairing. 
DMT-JEPA~\cite{dmtjepa} changes the context/target embeddings by aggregating features of semantically similar neighboring patches, mainly changing what latent target is predicted.
C-JEPA~\cite{cjepa} further augments the JEPA formulation with contrastive objectives to improve training stability and invariance alignment, while LeJEPA~\cite{lejepa} aims to provide a more principled and heuristic-free formulation.

\begin{figure}[t]
    \centering
    \includegraphics[width=\linewidth]{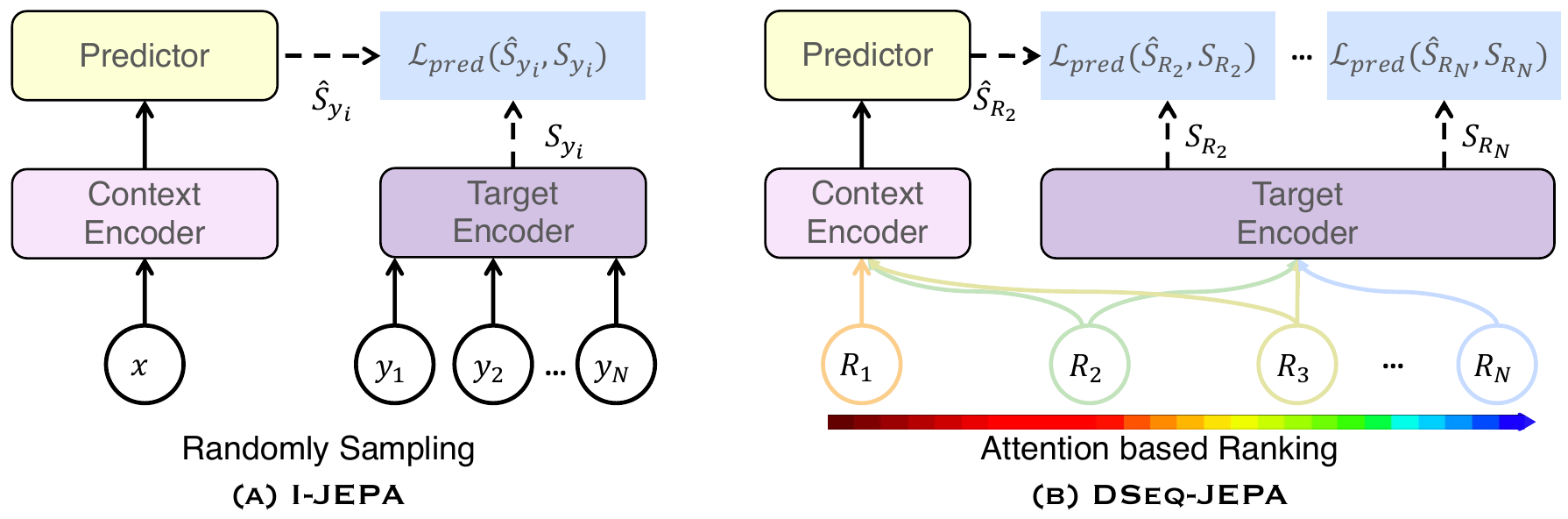}
    \caption{
    \textbf{(A) I-JEPA} predicts embeddings of target regions $(y_1, \ldots, y_N)$ from a context region $x$, using a predictor network. In contrast, \textbf{(B) \ours} uses an attention map to identify and select primary discriminative regions (\emph{where}), 
    and then performs sequential latent prediction from the most discriminative to the least (\emph{in what order}), through a predictor to predict the embeddings of the next discriminative regions $\{R_2$, \dots, $R_{N}\}$ based on the sequence of identified regions.
}
    \label{fig:fgjepavsijepa}
\end{figure}

Despite their strong abstraction capabilities, existing JEPA-based image pretraining methods treat target regions largely uniformly and predict them independently, without an explicit mechanism to determine which regions are most discriminative or in what order predictions should proceed. As shown in \Cref{fig:fgjepavsijepa}, \ours~ aims to improve JEPA-based representation learning
by introducing two complementary inductive biases: (i) \emph{discriminative region prioritization} (\emph{where to predict}), which biases prediction toward informative target regions instead of uniform target sampling, and (ii) \emph{ sequential latent prediction} (\emph{in what order to predict}), which replaces flat unordered prediction with a structured sequential predictive process. Compared with generative autoregressive models such as iGPT~\cite{igpt}, \ours~operates in the latent space rather than reconstructing image tokens, and uses a discriminative region ordering rather than a fixed token order.

\section{Methodology}
\label{sec:method}

\subsection{Overall Framework}
\begin{figure*}[t]
    \centering
    \includegraphics[width=\linewidth]{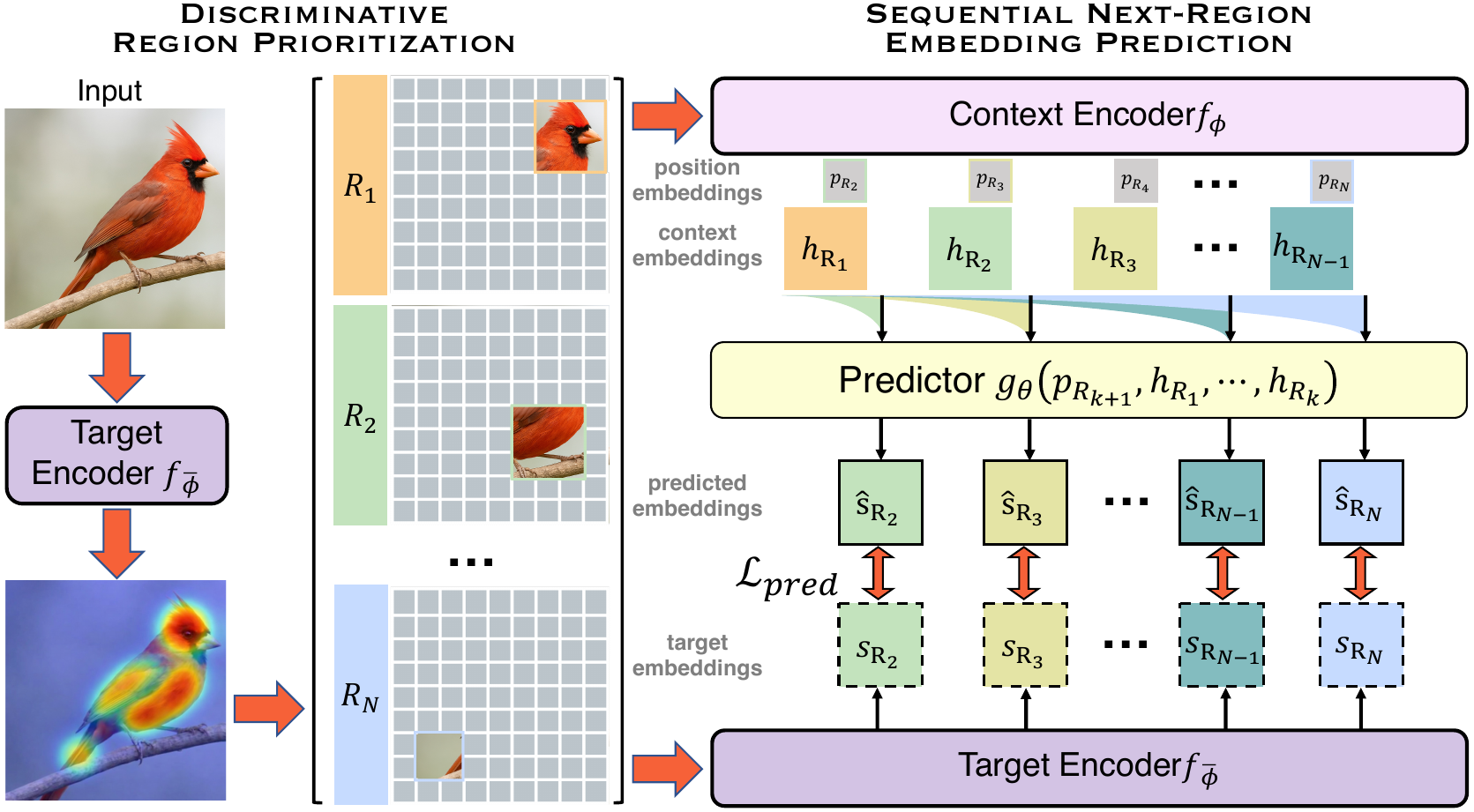}
    \caption{\textbf{Overview of \ours.}
     Given an input image, the target encoder $f_{\bar\phi}$ produces an attention-derived saliency map used to select and rank the top $N$ discriminative regions $\mathcal{R}=\{R_k\}_{k=1}^{N}$ (\emph{Discriminative Region Prioritization}). The context encoder $f_\phi$ then encodes these regions, and the predictor $g_\theta$ performs \emph{Sequential Next-Region Embedding Prediction}: at step $k$, it predicts the embedding of $R_{k+1}$ from the context features of previous regions and the positional token $\mathbf{p}_{R_{k+1}}$, and aligns it with the corresponding target-encoder embedding using latent prediction loss $\mathcal{L}_{\mathrm{pred}}$.}
    \label{fig:dseqjepa}
\end{figure*}

Our overall framework is illustrated in \Cref{fig:fgjepavsijepa}. Our goal is to enhance self-supervised learning by introducing a preferential strategy for predictive region selection and an explicit prediction order. To this end, we propose a framework that explicitly models both \emph{where} to predict and \emph{in what order} predictions should proceed. As shown in \Cref{fig:dseqjepa}, the framework comprises two key components: \emph{(1) Discriminative Region Prioritization}, which produces an attention-derived saliency map to identify informative regions (\emph{where} to predict) and to derive importance scores for ranking the selected regions (\emph{in what order} to predict); and \emph{(2) Sequential Next-Region Embedding Prediction}, which performs latent prediction sequentially from the most discriminative region to the least.

\subsection{Discriminative Region Prioritization}
\label{sec:drs}
Given an input image, we first generate an attention-derived saliency map $\mathbf{A}$ using the target encoder, which provides spatial guidance for locating informative visual content. Specifically, the saliency map is computed from the similarity between the auxiliary class token and patch embeddings at a selected transformer block, highlighting patches that contribute strongly to global image semantics. Formally, given an image $\mathbf{x} \in \mathbb{R}^{C\times H\times W}$, we feed it into the target encoder $f_{\bar{\phi}}$, obtaining feature embeddings $\mathbf{s}=f_{\bar{\phi}}(\mathbf{x}) \in \mathbb{R}^{D\times h\times w}$, where $D$ is the embedding dimension, and $h,w$ denote the number of vertical and horizontal patches, respectively. We compute a similarity map $\mathbf{A} \in \mathbb{R}^{h\times w}$ between the embeddings of the class token and all patches at the $l$-th transformer block. 
While we use this simple similarity-based attention for region selection, the formulation is general and can accommodate other attention mechanisms that provide semantically meaningful guidance (See \Cref{tab:atten_strategy}).
We then normalize $\mathbf{A}$ to $\tilde{\mathbf{A}}$ using Eq.~\eqref{eq:nor},
\begin{equation}
\label{eq:nor}
\tilde{\mathbf{A}}=
\frac{\mathbf{A}-\min(\mathbf{A})}{\max(\mathbf{A})-\min(\mathbf{A})},
\end{equation}
and apply Otsu’s method~\cite{Otsu1979}, which adaptively selects a data-driven threshold per image, to obtain a binary mask,
\begin{equation}
\mathbf{M}(i,j)=\mathds{1}\!\left[\tilde{\mathbf{A}}(i,j)\ge \tau\right].
\end{equation}
We apply connected-component labeling to $\mathbf{M}$ under a relaxed connectivity criterion (8-neighborhood) to obtain the candidate set $\{R_k\}$. After removing small components ($|R_k|<0.15hw$), each remaining region is assigned a discriminative score $\rho_k$ defined as the average normalized saliency response within the region (Eq.~\eqref{eq:att_score}),
\begin{equation}
\label{eq:att_score}
\rho_k=\frac{1}{|R_k|}\sum_{(i,j)\in R_k}\tilde{\mathbf{A}}(i,j).
\end{equation}
Unlike prior works~\cite{jepa,cjepa}, we employ irregular, non-overlapping masks for the extracted regions.
We rank regions by $\rho_k$, select the top $N-1$ regions as primary discriminative regions, and define the final region $R_N$ as the remaining image area to ensure complete image coverage:
\begin{equation}
\mathcal{R}=\{R_k\}_{k=1}^{N}, \qquad
R_N=\Omega\setminus\bigcup_{k=1}^{N-1}R_k,
\end{equation}
where $\Omega$ denotes the full image region.
This ensures complete coverage of the image without gaps, yielding more stable and discriminative region-based representation learning.

To improve stability when saliency maps are noisy early in pre-training, we adopt a probabilistic curriculum. Specifically, the probability $\lambda$ of applying discriminative region selection is linearly increased from $0$ to $1$ over pre-training epochs, so that each sample uses our discriminative region selection with probability $\lambda$ and random region sampling with probability $1-\lambda$. This progressive transition stabilizes optimization and gradually shifts the model from diffuse attention to structured saliency (see \Cref{alg:region_selection}).

\begin{algorithm}[t]
\caption{Discriminative Region Prioritization}
\label{alg:region_selection}
\KwIn{Saliency map $\mathbf{A}$, current epoch $t$, total pre-training epochs $T$}
\KwOut{Discriminative regions $\{R_k\}_{k=1}^{N}$}

Compute $\lambda \leftarrow \min\!\big(1,\max(0,\,t/T)\big)$\;

Normalize $\mathbf{A}$ to $\tilde{\mathbf{A}}$ via \Cref{eq:nor}\;
Obtain binary mask $\mathbf{M}$,  connected regions $\{R_{k}\}$\;
Compute discriminative score $\rho_{k}$ for each region via \Cref{eq:att_score}\;
Select top-$N$ regions $\mathcal{R}=\{R_k\}_{k=1}^{N}$\;

\For{$k=1$ \KwTo $N$}{
    Sample $b_k \sim \mathrm{Bernoulli}(\lambda)$; \\
    \eIf(){$b_k=1$}{
        $R_k$ := $R_k$;
    }{
    \DontPrintSemicolon
    \tcc*[l]{Randomly sample region with center$(u,v)$ and size$(w,h)$}
        $R_k$ := R(u,v,w,h);
    }
}
\Return $\{R_k\}_{k=1}^{N}$
\end{algorithm}

\subsection{Sequential Next-Region Embedding Prediction}
Given the ranked discriminative regions, \ours~predicts the next-region embeddings sequentially from the most discriminative region to the least.
Let $\mathcal{R}=\{R_k\}_{k=1}^{N}$ denote the selected regions sorted in descending order of discriminativeness. At each step $k \in \{1,\dots,N-1\}$, the predictor $g_\theta$ estimates the latent embedding of the $(k\!+\!1)$-th region, $\hat{\mathbf{s}}_{R_{k+1}}$, conditioned on the representations of the preceding discriminative regions from the context encoder, $\mathbf{h}_{R_1},\dots,\mathbf{h}_{R_k}$, together with the positional token $\mathbf{p}_{R_{k+1}}$ of the next target region:
\begin{equation}
\hat{\mathbf{s}}_{R_{k+1}} = g_\theta(\mathbf{p}_{R_{k+1}}, \mathbf{h}_{R_1}, \dots, \mathbf{h}_{R_k}).
\end{equation}
Compared with I-JEPA’s flat, independent target prediction, this formulation introduces ordered dependencies across regions and yields a structured and ordered sequential latent prediction process. 

\subsection{Training Objective}

To enforce the alignment between the predicted representation \(\hat{\mathbf{s}}_{R_{k+1}}\) and its ground-truth counterpart \(\mathbf{s}_{R_{k+1}}\), we adopt the smoothed \(\ell_1\) (Huber) loss.
Formally, the sequential latent prediction objective is defined as
\begin{gather}
\label{eq:lpred}
\mathcal{L}_{\mathrm{pred}}
= \frac{1}{N-1}\sum_{k=1}^{N-1} \sum_{j=1}^{D}
\psi\!\left(\bigl(\hat{\mathbf{s}}_{R_{k+1}}-\mathbf{s}_{R_{k+1}}\bigr)_j\right), \\
\label{eq:huber}
\psi(x)=
\begin{cases}
\tfrac{1}{2}x^2, & |x| < \delta,\\
\delta\left(|x|-\tfrac{1}{2}\delta\right), & |x|\ge \delta,
\end{cases}
\qquad \delta = 1.
\end{gather}
Here, $D$ denotes the embedding dimension, and $(\cdot)_j$ indexes the $j$-th feature dimension. This objective encourages stable and discriminative alignment between predicted and target embeddings, making sequential latent prediction smooth while remaining robust to outliers.


\section{Experiments}
\label{sec:experiments}

\subsection{Evaluation setup}
\label{sec:dataset}
To comprehensively evaluate the effectiveness of \ours~in learning discriminative and transferable visual representations, we pre-train all self-supervised models on ImageNet-1K~\cite{imagenet} using the standard protocol from I-JEPA \cite{ijepa} and C-JEPA \cite{cjepa}. We evaluate on five downstream benchmark groups: (1) image classification (ImageNet-1K~\cite{imagenet}); (2) fine-grained visual categorization (iNaturalist21~\cite{van2018inaturalist}, CUB-200-2011~\cite{cub}, and Stanford Cars~\cite{cars}); (3) object detection and instance segmentation (MS-COCO~\cite{coco}); (4) semantic segmentation (ADE20K~\cite{ade20k}); and (5) low-level reasoning (CLEVR~\cite{clevr}). We follow I-JEPA/C-JEPA pre-training and evaluation protocols whenever applicable for fair comparison. 

\subsection{Main Results}
\subsubsection{Image classification and fine-grained visual categorization.} 
\Cref{tab:linear_sota} reports results for linear probing and full fine-tuning on ImageNet-1K, together with fine-grained visual categorization (iNat21, CUB, and Cars). Linear probing on ImageNet-1K is a standard SSL evaluation protocol and directly reflects representation quality. Across \emph{ViT-B/16, ViT-L/16, and ViT-H/16} backbones, \ours~consistently improves over I-JEPA in linear probing 
(\(+1.1\), \(+0.8\), and \(+1.3\) Top-1 accuracy, respectively) and also improves full fine-tuning (\(+0.5\), \(+0.2\), and \(+0.7\), respectively). 
To further assess stability, we additionally repeat \ours~under the ImageNet ViT-B/16 setting with three random seeds, obtaining $73.8 \pm 0.4$ Top-1 accuracy, suggesting that the result is stable across runs.
The gains extend to fine-grained benchmarks, where \ours~improves over I-JEPA on all reported settings (e.g., ViT-B/16: \(+0.5\) on iNat21, \(+0.9\) on CUB, and \(+1.4\) on Cars; ViT-L/16: \(+0.7\), \(+1.2\), and \(+0.8\), respectively), yielding a consistent improvement in the FGVC average. Moreover, at larger scale, \ours~with ViT-H/16 substantially outperforms LeJEPA~\cite{lejepa} by \(+3.4\) points in ImageNet linear probing. These trends support our hypothesis that discriminative region prioritization and sequential next-region embedding prediction improve both global semantic separation and fine-grained cue discrimination.

Compared with C-JEPA, which augments I-JEPA with contrastive regularization, \ours~achieves comparable ImageNet performance under both linear probing and full fine-tuning, while consistently outperforming C-JEPA on fine-grained benchmarks (iNat21/CUB/Cars), indicating that discriminative sequential prediction learns more part-aware and discriminative representations. When adding the same contrastive regularization, \cours~ surpasses C-JEPA across settings, suggesting that contrastive alignment and discriminative sequential prediction are complementary.

\begin{table}[t]
  \centering
  \caption{\textbf{Linear probing and full fine-tuning evaluations.} We report \emph{Top-1 accuracy} on image classification (ImageNet) and fine-grained visual categorization (iNat21, CUB, Cars).  Average performance across all datasets is reported in the last column. 
  }
  \label{tab:linear_sota}
  \centering
  \resizebox{\linewidth}{!}{
  \begin{tabular}{llcllllll}
    \toprule
    \multirow{2}{*}{\textbf{Method}} & \multirow{2}{*}{\textbf{Arch.}} & \multirow{2}{*}{\textbf{Epochs}} & \multicolumn{2}{c}{\textbf{Image Classification}} & \multicolumn{3}{c}{\textbf{FGVC}} & \multirow{2}{*}{\textbf{Overall Avg.}} \\
    & & & Linear & Fine-tune & iNat21 & CUB & Cars & \\
    \midrule
    \multicolumn{9}{@{}l}{\emph{Methods without view data augmentations}} \\
    MAE~\cite{mae} & ViT-B/16 & 1600 & 68.0 \ \	\ \ \ \  & 83.6 \ \	\ \ \ \  & 33.5 \ \	\ \ \ \ & 59.8 \ \	\ \ \ \	& 60.2 \ \	\ \ \ \	&  61.0 \ \	\ \ \ \\
    I-JEPA~\cite{ijepa} & ViT-B/16 & 600 & 72.4 \ \	\ \ \ \ &83.5 \ \	\ \ \ \ & 35.9 \ \	\ \ \ \ & 65.3 \ \	\ \ \ \	& 65.9 \ \	\ \ \ \	&  64.6 \ \	\ \ \ \ \\
    \rowcolor{myblue}
    \textbf{\ours} &  ViT-B/16 & 600 & \textbf{73.5}\textsubscript{ {\textcolor{mygreen}{+1.1}}} & \textbf{84.0}\textsubscript{ {\textcolor{mygreen}{+0.5}}} & \textbf{36.4}\textsubscript{ {\textcolor{mygreen}{+0.5}}}	& \textbf{66.2}\textsubscript{ {\textcolor{mygreen}{+0.9}}}	& \textbf{67.3}\textsubscript{ {\textcolor{mygreen}{+1.4}}}	&  \textbf{65.5}\textsubscript{ {\textcolor{mygreen}{+0.9}}} \\
    C-JEPA~\cite{cjepa} & ViT-B/16 & 600 & 73.5 \ \	\ \ \ \ &84.2 \ \	\ \ \ \	&36.2 \ \	\ \ \ \	& 64.9 \ \	\ \ \ \	& 66.1 \ \	\ \ \ \ & 65.0 \ \	\ \ \ \ \\
    \rowcolor{myblue}
    \textbf{\cours} & ViT-B/16 & 600 & \textbf{73.8}\textsubscript{ {\textcolor{mygreen}{+0.3}}}& \textbf{84.3}\textsubscript{ {\textcolor{mygreen}{+0.1}}} & \textbf{36.6}\textsubscript{ {\textcolor{mygreen}{+0.4}}} & \textbf{66.5}\textsubscript{ {\textcolor{mygreen}{+1.6}}}& \textbf{67.4}\textsubscript{ {\textcolor{mygreen}{+1.3}}} & \textbf{65.7}\textsubscript{ {\textcolor{mygreen}{+0.7}}} \\
    NEPA~\cite{nepa} & ViT-B/14 & 1600 &  \ \ -	\ \ \ \	& 83.8 \ \	\ \ \ \	&  \ \ -	\ \ \ \	&  \ \ -	\ \ \ \	&  \ \ -	\ \ \ \	&  \ \ -	\ \ \ \	\\
    \midrule
    MAE~\cite{mae} & ViT-L/16 & 1600 & 76.0 \ \	\ \ \ \	& 85.9 \ \	\ \ \ \	& 37.1 \ \	\ \ \ \	& 63.4 \ \	\ \ \ \	& 61.9 \ \	\ \ \ \	& - \ \	\ \ \ \ \\
    I-JEPA~\cite{ijepa} & ViT-L/16 & 600 & 77.1 \ \	\ \ \ \ &86.6 \ \	\ \ \ \	& 38.8 \ \	\ \ \ \	& 66.9 \ \	\ \ \ \	& 68.1 \ \	\ \ \ \	& 67.5 \ \	\ \ \ \ \\
    \rowcolor{myblue}
    \textbf{\ours} & ViT-L/16 & 600 & \textbf{77.9}\textsubscript{ {\textcolor{mygreen}{+0.8}}}	& \textbf{86.8}\textsubscript{ {\textcolor{mygreen}{+0.2}}}& \textbf{39.5}\textsubscript{ {\textcolor{mygreen}{+0.7}}}	& \textbf{68.1}\textsubscript{ {\textcolor{mygreen}{+1.2}}}	& \textbf{68.9}\textsubscript{ {\textcolor{mygreen}{+0.8}}}	&  \textbf{68.2}\textsubscript{ {\textcolor{mygreen}{+0.8}}} \\
    C-JEPA~\cite{cjepa} & ViT-L/16 & 600 & 78.0 \ \	\ \ \ \ &86.9 \ \	\ \ \ \	& 39.0 \ \	\ \ \ \	& 65.8 \ \	\ \ \ \ & 67.7 \ \	\ \ \ \	&  67.5 \ \	\ \ \ \ \\
    \rowcolor{myblue}
    \textbf{\cours} & ViT-L/16 & 600 & \textbf{78.4}\textsubscript{ {\textcolor{mygreen}{+0.4}}} & \textbf{87.2}\textsubscript{ {\textcolor{mygreen}{+0.3}}} & \textbf{39.7}\textsubscript{ {\textcolor{mygreen}{+0.7}}}	& \textbf{68.3}\textsubscript{ {\textcolor{mygreen}{+2.5}}} & \textbf{68.8}\textsubscript{ {\textcolor{mygreen}{+1.1}}}	&  \textbf{68.5}\textsubscript{ {\textcolor{mygreen}{+1.0}}} \\
    NEPA~\cite{nepa} & ViT-L/14 & 800 &  \ \ -	\ \ \ \	& 85.3 \ \	\ \ \ \	&  \ \ -	\ \ \ \	&  \ \ -	\ \ \ \	&  \ \ -	\ \ \ \	&  \ \ -	\ \ \ \	\\
    \midrule
    MAE~\cite{mae} & ViT-H/14 & 1600 & 77.2 \ \	\ \ \ \ &  \ \ -	\ \ \ \	&  \ \ -	\ \ \ \	&  \ \ -	\ \ \ \	&  \ \ -	\ \ \ \	&  \ \ -	\ \ \ \	\\
     I-JEPA~\cite{ijepa} & ViT-H/16$_{448}$ & 300 & 81.1 \ \	\ \ \ \ &  87.1	&  38.9	&  67.2	&  67.7	&  68.4	\\
    \rowcolor{myblue}
    \textbf{\ours} & ViT-H/16$_{448}$ & 300 & \textbf{82.4}\textsubscript{ {\textcolor{mygreen}{+1.3}}}	& \textbf{87.8}\textsubscript{ {\textcolor{mygreen}{+0.7}}}	& \textbf{39.3}\textsubscript{ {\textcolor{mygreen}{+0.4}}}	& \textbf{68.9}\textsubscript{ {\textcolor{mygreen}{+1.7}}}	& \textbf{70.1}\textsubscript{ {\textcolor{mygreen}{+2.4}}}	&\textbf{69.7}\textsubscript{ {\textcolor{mygreen}{+1.5}}}	\\
    \toprule
    \multicolumn{9}{@{}l}{\emph{Methods with view data augmentations}} \\
    LeJEPA~\cite{lejepa} & ViT-H/14 & 100 & 79.0 \ \	\ \ \ \ &  \ \ -	\ \ \ \	&  \ \ -	\ \ \ \	&  \ \ -	\ \ \ \	&  \ \ -	\ \ \ \	&  \ \ -	\ \ \ \	\\
    DINO~\cite{dino} & ViT-B/16 & 1600 & 78.2 \ \	\ \ \ \ & 82.8 \ \	\ \ \ \ & 30.2 \ \	\ \ \ \ & 33.6 \ \	\ \ \ \	& 27.0 \ \	\ \ \ \	&  50.4 \ \	\ \ \ \ \\ 
    iBOT~\cite{ibot} & ViT-B/16 & 1600 & 79.5 \ \	\ \ \ \ & 84.0 \ \	\ \ \ \ & 38.9 \ \	\ \ \ \ & 67.7 \ \	\ \ \ \	& 68.1 \ \	\ \ \ \	&  67.6 \ \	\ \ \ \ \\ 
    \bottomrule
  \end{tabular}
  }
\end{table}

Relative to representative SSL baselines (MAE, DINO and iBOT), \ours~remains competitive under a simpler JEPA-style single-view pre-training recipe, compared to reconstruction- and multi-view-based baselines. On ViT-B/16, DINO and iBOT (with multi-view augmentations) achieve higher ImageNet linear-probing accuracy, whereas \ours~remains competitive and yields strong fine-tuning performance despite relying on a simpler single-view recipe. At larger scale, \ours~with ViT-H attains the highest ImageNet accuracy under both linear probing and full fine-tuning. In addition, \ours~is training-efficient: it uses a \emph{single-view} recipe (vs.\ DINO/iBOT's \emph{multi-view augmentation}) and runs for \emph{600/300} epochs, whereas MAE/DINO/iBOT require \emph{1600} epochs. This efficiency advantage of JEPA-style single-view pre-training is consistent with observations from I-JEPA, where a large ViT-H/14 I-JEPA model requires less compute than a smaller iBOT ViT-S/16 model~\cite{jepa}.

Additionally, we compare with NEPA~\cite{nepa}, a recent sequential embedding-prediction baseline. Although the training configurations are not strictly matched, both \ours~and \cours~achieve higher ImageNet fine-tuning accuracy on both ViT-B (84.0/84.3 vs.\ 83.8) and ViT-L (86.8/87.2 vs.\ 85.3), suggesting that semantically grounded next-region prediction is more effective than generic sequential embedding prediction.
\begin{table}[t]
  \caption{\textbf{Performance on detection and segmentation tasks.} We report \emph{AP\textsuperscript{box}}, \emph{AP\textsuperscript{mask}} on MS-COCO, and \emph{mIoU} on ADE20K.
  }
  \label{tab:detection}
  \centering
  \begin{tabular}{>{\arraybackslash}m{3cm}>{\centering\arraybackslash}m{1.8cm}>{\centering\arraybackslash}m{1.8cm}>{\raggedright\arraybackslash}m{1.6cm}>{\raggedright\arraybackslash}m{1.6cm}>{\raggedright\arraybackslash}m{1.6cm}}
    \toprule
    \textbf{Method} & \textbf{Arch.} & \textbf{Epochs} & \textbf{AP\textsuperscript{box}} & \textbf{AP\textsuperscript{mask}} & \textbf{mIoU}\\
    \midrule
    \multicolumn{5}{@{}l}{\emph{Methods without view data augmentations}} \\
    MAE~\cite{mae} & ViT-B/16 & 1600 & 50.3 \ \	\ \ \ \	& 44.9 \ \	\ \ \ \	& 48.1 \ \	\ \ \ \	\\
    I-JEPA~\cite{ijepa} & ViT-B/16 & 600 & 49.9 \ \	\ \ \ \ & 44.5 \ \	\ \ \ \ & 47.6 \ \	\ \ \ \ \\
    \rowcolor{myblue}
    \textbf{\ours} & ViT-B/16 & 600 & \textbf{50.5} \textsubscript{{\color{mygreen}+0.6}} & \textbf{45.0}\textsubscript{ {\color{mygreen}+0.5}} & \textbf{48.1}\textsubscript{ {\color{mygreen}+0.5}} \\ 
    C-JEPA~\cite{cjepa} & ViT-B/16 & 600 & 50.7 \ \	\ \ \ \ & 45.3 \ \	\ \ \ \ & 48.7 \ \	\ \ \ \ \\
    \rowcolor{myblue}
    \textbf{\cours} & ViT-B/16 & 600 & \textbf{50.9}\textsubscript{ {\color{mygreen}+0.2}} & \textbf{45.7}\textsubscript{ {\color{mygreen}+0.4}} & \textbf{48.9}\textsubscript{ {\color{mygreen}+0.2}} \\
    NEPA~\cite{nepa} & ViT-B/14 & 1600 &  \ \ -	\ \ \ \	&  \ \ -	\ \ \ \	& 48.3 \ \	\ \ \ \	\\
    \toprule    
    \multicolumn{5}{@{}l}{\emph{Methods with view data augmentations}} \\
    DINO~\cite{dino} & ViT-B/16 & 1600 & 50.1 \ \	\ \ \ \	& 43.4 \ \	\ \ \ \	& 46.8 \ \	\ \ \ \	\\
    iBOT~\cite{ibot} & ViT-B/16 & 1600 & 51.2 \ \	\ \ \ \	& 44.2 \ \	\ \ \ \	& 50.0 \ \	\ \ \ \	\\
    \bottomrule
  \end{tabular}
\end{table}

\subsubsection{Detection and Segmentation.}
To further evaluate transferability to dense prediction, we evaluate object detection and instance segmentation on MS-COCO~\cite{coco} and semantic segmentation on ADE20K~\cite{ade20k}. As shown in \Cref{tab:detection}, \ours~consistently improves over I-JEPA across all reported metrics (\(+0.6\) AP\textsuperscript{box}, \(+0.5\) AP\textsuperscript{mask}, and \(+0.5\) mIoU), while \cours~achieves the best overall performance among JEPA variants, surpassing both I-JEPA and C-JEPA. \cours~also exceeds NEPA on ADE20K mIoU (48.9 vs.\ 48.3), suggesting that discriminative ordering benefits not only sequential prediction but also spatial generalization. Despite using a lighter single-view 600-epoch recipe, \ours/\cours~ outperform MAE and DINO across all reported dense-prediction metrics, and \cours~achieves higher AP\textsuperscript{mask} than iBOT. Overall, these results indicate that discriminative sequential prediction promotes more spatially structured and semantically coherent representations that transfer effectively beyond classification to pixel-level prediction.

\subsubsection{Low-level Reasoning.}
As shown in \Cref{tab:clevr}, on low-level reasoning tasks, \ours~achieves 86.4 on Clevr/Count and 71.5 on Clevr/Dist~\cite{clevr} with ViT-L/16, improving over I-JEPA by \(+0.8\) and \(+0.3\), respectively. With contrastive regularization, \cours~further improves to 87.1/71.9, surpassing C-JEPA on both tasks. The fact that improvements persist on these low-level metrics indicates that \ours~does not merely enhance semantic categorization, but also strengthens geometric and structural priors in the learned representation.

\begin{table}[t]
  \caption{\textbf{Performance on 
  low-level tasks.} 
  We report \emph{Clevr/Count} (object counting) and \emph{Clevr/Dist} (distance estimation) with a linear probe on frozen ViT-L/16 features, measured by \emph{Top-1 accuracy}.
  }
  \label{tab:clevr}
  \centering
  \begin{tabularx}{\linewidth}{@{\extracolsep{\fill}}lcccc}
    \toprule
    \textbf{Metric} & I-JEPA & \cellcolor {myblue}\textbf{\ours}  & C-JEPA & \cellcolor{myblue}\textbf{\cours} \\
    \midrule
    \textbf{Clevr/Count} &  85.6 & \cellcolor{myblue}\textbf{86.4}\textsubscript{ {\color{mygreen}+0.8}} & 86.8 & \cellcolor{myblue}\textbf{87.1}\textsubscript{ {\color{mygreen}+0.3}} \\
    \textbf{Clevr/Dist} & 71.2 & \cellcolor{myblue}\textbf{71.5}\textsubscript{ {\color{mygreen}+0.3}} &71.6 &\cellcolor{myblue}\textbf{71.9}\textsubscript{ {\color{mygreen}+0.3}}\\
    \bottomrule
  \end{tabularx}
\end{table}
\begin{table}[t]
  \caption{\textbf{Effects of region generation and prediction strategy (ViT-B/16).} 
  }
  \label{tab:ab_classification}
  \centering
  \resizebox{\linewidth}{!}{
  \begin{tabular}{cc>{\centering\arraybackslash}m{1.6cm}>{\centering\arraybackslash}m{1.6cm}cc}
    \toprule
    \multicolumn{2}{c}{\textbf{Region Generation}} & \multicolumn{2}{c}{\textbf{Prediction Strategy}} & \multirow{2}{*}{\textbf{ImageNet}} & \multirow{2}{*}{\textbf{iNat21}} \\
    \emph{Uniform sampling} & \emph{Discriminative selection} & \emph{Flat} & \emph{Sequential} & & \\
    \midrule
    {\color{Green}\faCheckSquare} & & {\color{Green}\faCheckSquare} & & 72.4 & 35.9\\
    {\color{Green}\faCheckSquare} & &  & {\color{Green}\faCheckSquare} &  72.3	& 34.9 \\
    & {\color{Green}\faCheckSquare} & {\color{Green}\faCheckSquare} &  & 72.0	& 35.7  \\
    \rowcolor{myblue}
    & {\color{Green}\faCheckSquare} &  & {\color{Green}\faCheckSquare} & \textbf{73.5}  & \textbf{36.4}\\
    \bottomrule
  \end{tabular}
  }
\end{table}
\subsection{Ablation Study}

\subsubsection{Do discriminative selection and sequential prediction act synergistically?}
We ablate the two key components of \ours---\emph{discriminative region prioritization} and \emph{sequential next-region embedding prediction}---by varying region generation (uniform vs.\ discriminative) and prediction strategy (flat vs.\ sequential). \Cref{tab:ab_classification} reports ImageNet and iNat21 linear-probing accuracy with a ViT-B/16 backbone. The baseline configuration (uniform sampling + flat prediction) corresponds to I-JEPA, which treats all regions equally and predicts them in parallel. Enabling either component in isolation degrades performance: sequential prediction without an informative order (uniform + sequential) breaks permutation symmetry arbitrarily and introduces noisy supervision, while discriminative selection without sequential conditioning (selective + flat) collapses inter-region dependencies into independent targets. In contrast, combining both components (selective + sequential) yields the best results. These findings show that the modules are not independently effective but \emph{synergistic}: discriminative region selection supplies a semantically meaningful trajectory (where), and sequential prediction exploits this trajectory to capture region-to-region dependencies (in what order).

\begin{table}[t]
\centering
\caption{\textbf{Effects of different prediction orders (ViT-B/16).}}
\label{tab:seq-order}
\begin{tabularx}{\linewidth}{@{\extracolsep{\fill}}ccccccc}
\toprule
\textbf{Order scheme} & Flat & Random & Spatial & Inverse & Truncating & \textbf{DSeq (ours)} \\
\midrule
\textbf{ImageNet} & 72.0 & 71.7 & 72.7 & 71.3 & 73.0 & \textbf{73.5} \\
\bottomrule
\end{tabularx}
\end{table}

\subsubsection{How important is the prediction order?}
We next isolate the effect of prediction order while keeping the same set of discriminative regions. \Cref{tab:seq-order} compares several ordering schemes on ImageNet with ViT-B/16: \emph{Flat} (I-JEPA-style, no autoregression), \emph{Random} (random region order), \emph{Spatial} (regions sorted by center coordinates in row-major order), \emph{Truncating} (predict only up to Top-3 regions), and our \emph{DSeq} order (\ours).
We observe that random ordering harms performance, indicating that autoregression without a meaningful trajectory forces the model to learn from mixed-difficulty signals. 
A simple spatial order brings a measurable improvement (+0.7 over flat), suggesting that purely geometric structure provides limited benefit. 
The inverse order only reaches 71.3, showing that the direction of the prediction trajectory matters. 
Our discriminative sequential order achieves the highest accuracy (+1.5 over flat), showing that encoding a semantically grounded prediction trajectory over regions is crucial for reaping the benefits of sequential prediction. 

\begin{figure}[t]
\begin{minipage}[t]{0.47\linewidth}
  \vspace{0pt} 
  \centering
  \includegraphics[width=\linewidth]{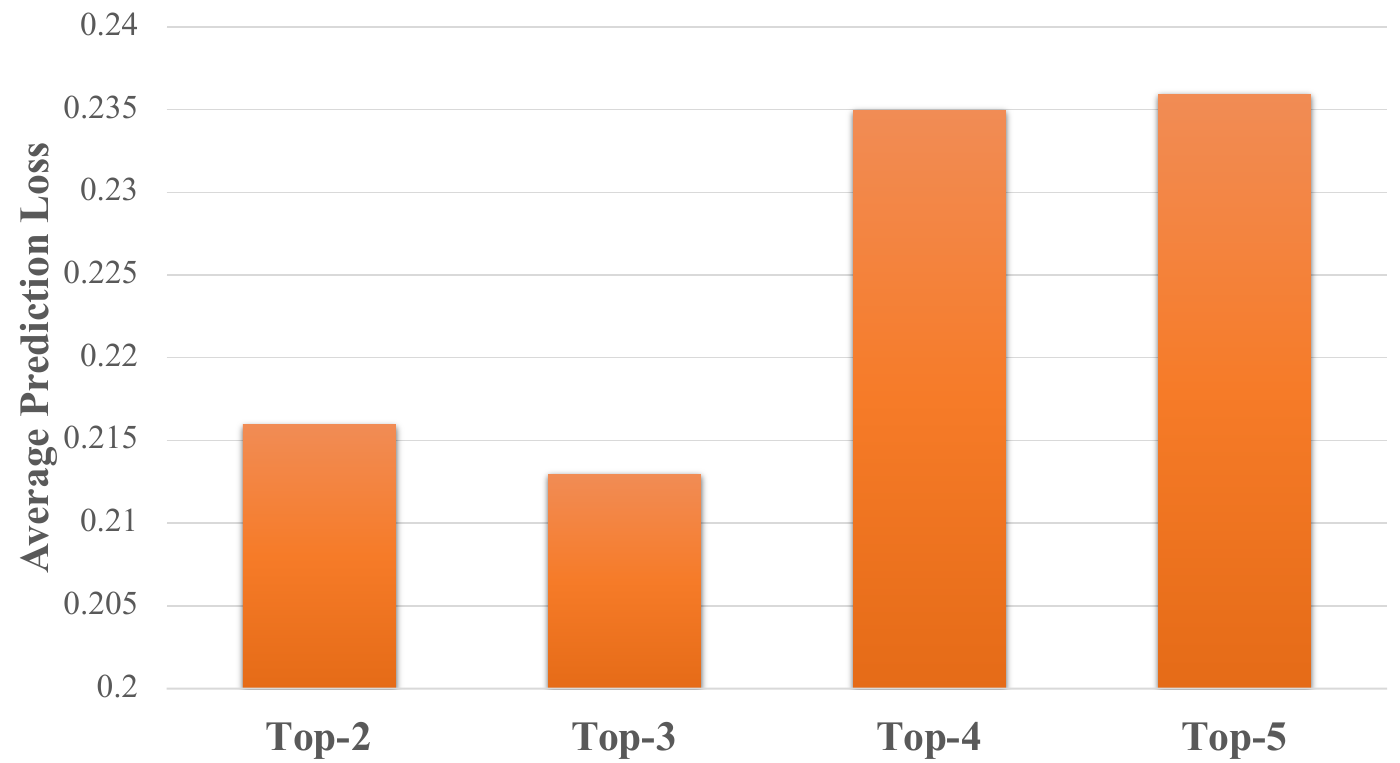}
  \caption{\textbf{Per-step prediction difficulty} (ViT-B/16 checkpoint at epoch 450). 
  }
  \label{fig:per_step}
\end{minipage}
\begin{minipage}[t]{0.5\linewidth}
  \vspace{0pt} 
  \centering
  \includegraphics[width=\linewidth]{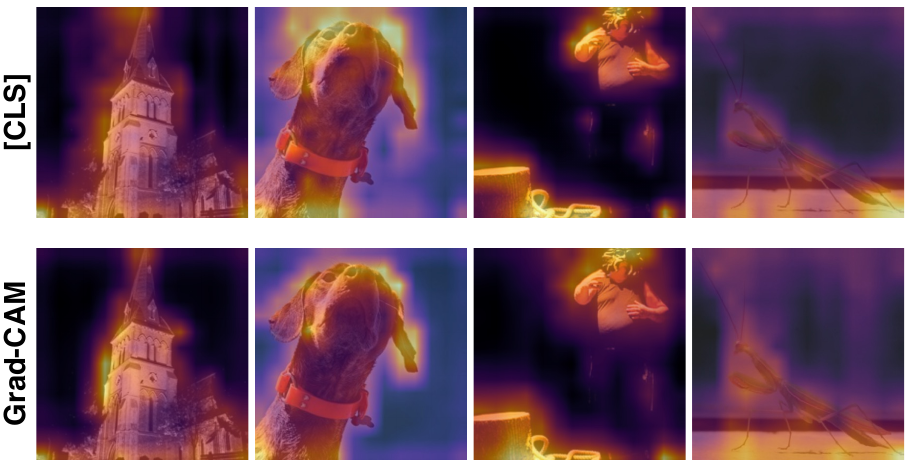}
  \caption{\textbf{\texttt{[CLS]}-based saliency vs. Grad-CAM.} 
  }
  \label{fig:cls_gradcam}
\end{minipage}\hfill
\end{figure}

Furthermore, the step-truncation diagnostic shows that predicting only up to the Top-3 regions (\emph{Truncating}) yields 73.0 vs.\ 73.5 with the full chain (Top-5), indicating that later, lower-saliency regions still provide complementary supervision beyond the most salient cues. Together, these ablations support our design choice of using a discriminative, full-length prediction order rather than generic or truncated autoregression.
Viewed from a learning-dynamics perspective, the discriminative order induces an implicit easy-to-hard curriculum: early steps focus on stable, highly informative regions, and later steps progressively integrate more context-dependent cues. To quantify this effect, we analyze the per-step prediction loss along the next-region prediction sequence using \ours~ViT-B checkpoint at epoch 450. We sample 10{,}000 ImageNet images and compute the average loss for each Top-$k$ region prediction. As shown in \Cref{fig:per_step}, early steps (Top-2 and Top-3) have lower prediction error, whereas later steps (Top-4 and Top-5) are clearly harder with noticeably higher loss. This trend suggests that the model first learns to predict stable, highly informative regions and then progressively integrates weaker or more context-dependent cues, and the sequential predictor exploits this ordered trajectory to shape the learned representations.

\subsubsection{Effect of the \texttt{[CLS]} Token.}
\label{sec:exp_cls}
To keep the architecture comparable to I-JEPA, we introduce an auxiliary \texttt{[CLS]} token that is used only as an anchor to compute saliency maps, without adding extra layers or parameters. As shown in \Cref{tab:clstoken}, adding this token does not improve I-JEPA’s performance, indicating that it provides no representational gain by itself. Thus, the improvements of \ours~stem from the proposed discriminative region prioritization and sequential next-region prediction, rather than from any architectural modification.

\subsubsection{Robustness to Attention Proxies.}
\Cref{tab:atten_strategy} assesses whether \ours~depends on a specific saliency estimator by replacing the \texttt{[CLS]}–based proxy with a label-free Grad-CAM–style proxy~\cite{gradcam}. This variant attains 73.4 ImageNet linear-probing accuracy (vs.\ \texttt{[CLS]}–based: 73.5), indicating that our gains are robust to the choice of attention proxy. 
\Cref{fig:cls_gradcam} shows that our \texttt{[CLS]}-based saliency closely overlaps with the Grad-CAM proxy, achieving a Top-20 patch IoU of $0.41$ while being simpler and gradient-free.
Together with the prediction-order ablations in \Cref{tab:seq-order}, this suggests that performance is driven not by a particular saliency mechanism, but by exploiting a semantically meaningful prediction trajectory: arbitrary orderings can hurt performance, whereas orderings that better correlate with discriminative content consistently yield gains, and the proxy simply provides a practical way to approximate such an ordering prior.

\begin{figure}[t]
\begin{minipage}[t]{0.4\linewidth}
  \vspace{0pt} 
  \centering

  \captionof{table}{\textbf{Effect of \texttt{[CLS]} token.}}
  \label{tab:clstoken}
  \vspace{0.2em}
  \begin{tabular}{ccc}
    \toprule
    \textbf{Method} & I-JEPA & +\texttt{[CLS]}\\
    \midrule
    \textbf{ImageNet} & 72.4 & 72.4 \\
    \bottomrule
  \end{tabular}
\end{minipage}\hfill
\begin{minipage}[t]{0.55\linewidth}
  \vspace{0pt} 
  \centering
  \captionof{table}{\textbf{Robustness to attention proxies.}}
  \label{tab:atten_strategy}
  \vspace{0.2em}
  \begin{tabular}{ccc}
    \toprule
    \textbf{Method} & Grad-CAM-based & \texttt{[CLS]}-based\\
    \midrule
    \textbf{ImageNet} & 73.4 & 73.5 \\
    \bottomrule
  \end{tabular}
\end{minipage}
\end{figure}





\subsubsection{Sensitivity to the Number of Regions $N$.}
We fix the region number to $N=5$, following the I-JEPA setting. To assess sensitivity to this choice, we vary $N \in \{3,5,7\}$ for \ours~with ViT-B/16 and pre-train on ImageNet under the same protocol.
The resulting ImageNet linear-probing accuracies are $72.9$, $73.5$, and $73.4$, respectively. 
The overall performance fluctuates only mildly, indicating that \ours~is not sensitive to the exact number of discriminative regions. 
Using too few regions ($N=3$) slightly under-utilizes contextual information, while increasing to $N=7$ brings no further benefit.

\vspace{-0.05in}
\subsection{Pre-training Overhead and Inference Efficiency}
\Cref{tab:cost} reports pre-training and inference costs on ImageNet with ViT-B/16 under the same hardware setup. Notably, compared with I-JEPA/C-JEPA, \ours's inference complexity remains essentially unchanged (86.6M parameters; 17.7 vs.\ 17.8 GFLOPs/image), since discriminative sequential prediction is only used during pre-training. The modest added computation introduced by \ours~is confined to pre-training: it adds about +2.0 hours wall-clock time (24.2/24.5~$\to$~26.5h), increases total training compute from 96.4/98.0 to 111.6 GFLOPs, and raises peak memory usage from 31.5 to 38.2\,GB due to sequential region tokens and causal masking. \cours~follows the same pattern.



\begin{table}[t]
  \caption{\textbf{Pre-training overhead and inference efficiency.} }
  \label{tab:cost}
  \centering
  \resizebox{\linewidth}{!}{
  \begin{tabular}{lccccc}
    \toprule
    \multirow{2}{*}{\textbf{Method}} & \multicolumn{3}{c}{\textbf{Pre-training}} & \multicolumn{2}{c}{\textbf{Inference}} \\
    & Time(h) & Mem(GB) & Total GFLOPs & \#Params(M) & GFLOPs/image \\
    \midrule
    I-JEPA & 24.2 & 31.5 & 96.4 & 86.6 & 17.7\\
    C-JEPA & 24.5 & 31.5 & 98.0 & 86.6 & 17.7 \\
    \rowcolor{myblue}
    \textbf{\ours} & 26.5 & 38.2 & 111.6 & 86.6 & 17.8\\
    \rowcolor{myblue}
    \textbf{\cours} & 26.8 & 38.2 & 117.2 & 86.6 & 17.8 \\
    \bottomrule
  \end{tabular}
  }
\end{table}

\vspace{-0.05in}
\subsection{Visualization Analysis}

\subsubsection{Qualitative Comparison with I-JEPA.} 
As shown in \Cref{fig:discrimination}, I-JEPA yields relatively diffuse attention maps and rectangular target masks that often cover multiple, partially overlapping regions. In contrast, \ours~produces compact, semantically meaningful regions ordered by discriminativeness, progressively focusing on key object parts (e.g., the bird’s beak and forehead) and yielding clearer region correspondences and more interpretable predictions.

\subsubsection{Emergence of Semantic Structure during Pre-training.}
\begin{figure}[t]
\centering
\begin{minipage}{0.57\linewidth}
  \centering
  \includegraphics[width=\linewidth]{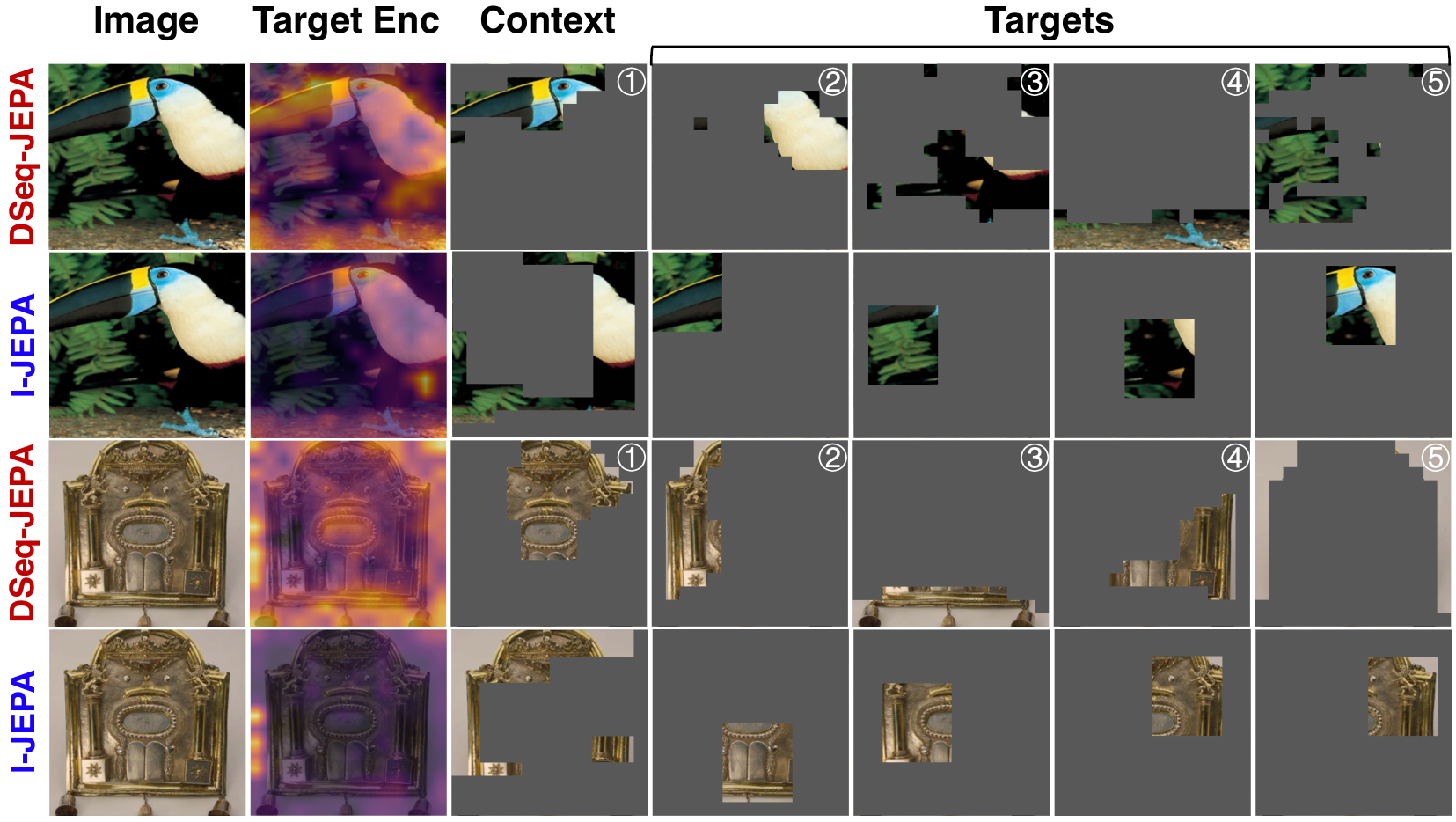}
  \caption{\textbf{Qualitative visualization of learned attention and selected context/target regions using ViT-B/16 model.} 
    For \ours, the numbered regions (\Circled{1}–\Circled{5}) correspond to discriminative regions ordered by their estimated importance.}
  \label{fig:discrimination}
\end{minipage}\hfill
\begin{minipage}{0.4\linewidth}
  \centering
  \includegraphics[width=\linewidth]{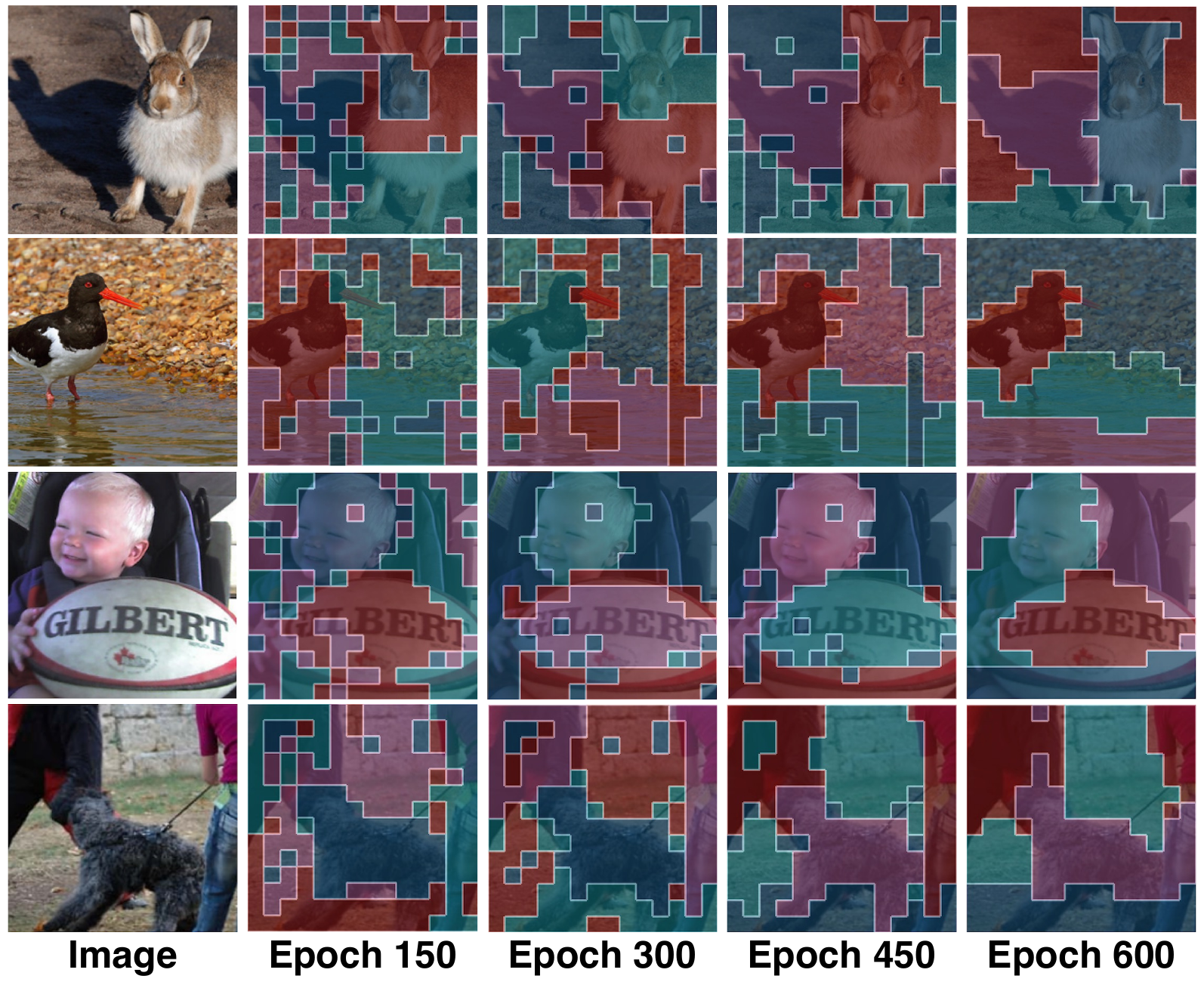}
  \caption{\textbf{Evolution of patch-level clustering during pre-training.} From left to right: input image and 4-cluster results from ViT-B/16 checkpoints at 150, 300, 450, and 600 epochs.}
  \label{fig:patch_cluster}
\end{minipage}
\end{figure}

\Cref{fig:patch_cluster} visualizes the evolution of patch-level clustering as pre-training proceeds. Early in training, clusters are fragmented and noisy; over time, they self-organize into coherent, object-aligned groups, even in the absence of labels. This illustrates how \ours’s representations self-organize during training: patches with similar semantics gradually move closer in the learned embedding space.
This progression indicates that \ours~gradually induces fine-grained semantic structure via its attention-guided, sequential prediction mechanism.


\section{Conclusion}
\label{sec:conclusion}
We introduced \ours, a JEPA-based SSL framework that augments latent prediction with \emph{discriminative region prioritization} and \emph{sequential next-region embedding prediction}. By explicitly modeling both \emph{where} to attend and \emph{in what order} to predict, \ours~progressively activates primary cues and then integrates secondary contexts, yielding more structured and fine-grained representations. 
Comprehensive evaluations across multiple benchmarks demonstrate consistent gains over JEPAs, confirming \ours's effectiveness and generality.
In future work, we plan to extend these ideas to vision–language pre-training, enabling more selective visual grounding and structured cross-modal alignment.


\section*{Acknowledgements}
This work was funded, in part, by the UBC Data Science Institute, UBC AI \& Health Network, Mitacs, Vector Institute for AI, Canada CIFAR AI Chair, NSERC Canada Research Chair (CRC), NSERC Discovery Grant, NSF CAREER Award No. 2441060 and NSF (2330423) \& NSERC (585136) Global Center on AI and Biodiversity Change. Resources used in preparing this research were provided, in part, by the Province of Ontario, the Government of Canada through CIFAR, the Digital Research Alliance of Canada, companies\footnote{ https://vectorinstitute.ai/\#partners} sponsoring the Vector Institute. Additional hardware support was provided by John R. Evans Leaders Fund CFI grant and Digital Research Alliance of Canada under the Resource Allocation Competition award. This work also benefited from JSPS KAKENHI Grant-in-Aid for Scientific Research (B) and (C) under Grant 26K02879 and 23K11164, and from French state funds managed by the National Research Agency via the IHU Strasbourg (ANR-10-IAHU-0002) and the ENACT AI Cluster (ANR-23-IACL-0004).
\bibliographystyle{unsrt} 
{
\bibliography{reference/ref}
}

\appendix
\section{Supplementary Material}

\subsection{Implementation Details}
\label{sup:implementation}
\subsubsection{(1) Model Architecture} \mbox{}\\
%

\noindent Following I-JEPA~\cite{ijepa}, our model consists of three modules: a context encoder $f_{\phi}$, a target encoder $f_{\bar{\phi}}$, and a predictor $g_{\theta}$, all built upon the Vision Transformer (ViT) \cite{vit}. The context and target encoders follow the standard ViT design, while the predictor is implemented as a lightweight ViT variant.
We evaluate \ours~with ViT-B/16, ViT-L/16, and ViT-H/16 backbones to demonstrate its effectiveness and scalability. 

While the overall architecture closely follows I-JEPA, we append an auxiliary class token to both the context and target encoders, following MAE~\cite{mae}, to maintain compatibility with standard ViT-based downstream tasks. 
This token acts as a stable semantic reference that slightly improves the reliability of attention-based region selection without altering the predictive formulation or training objective.

\subsubsection{(2) Pre-training}
\paragraph{Optimization.}
We follow the same optimization setting as I-JEPA~\cite{ijepa} and C-JEPA~\cite{cjepa}, using the AdamW optimizer with cosine learning rate scheduling and linear warmup. 
The learning rate starts from $1e$-$4$ and linearly increases to $1e$-$3$ during the first 15 epochs, and then decays to $1e$-$6$ following a cosine schedule. 
The weight decay is also scheduled with cosine annealing, increasing from $4e$-$2$ to $4e$-$1$. 
The target encoder is initialized with the context encoder weights and updated via exponential moving average (EMA) with decay factors gradually increasing from $0.996$ to $1.0$. 

\paragraph{Training Protocol.}
We pre-train \ours~for $600$ epochs with input images resized to $224\times224$ for ViT-B/16 and ViT-L/16. 
For the larger ViT-H/16 model, we train for $300$ epochs with a higher input resolution of $448\times448$. The global batch size was set to $2048$. Data augmentation techniques included random resized cropping with a random scale within $[0.3, 1.0]$. Following I-JEPA, we do not use hand-crafted data augmentations such as horizontal flipping, Gaussian blur, and color distortion. 
The saliency map is extracted from the $10$th layer, and regions ($|R_k|<0.15hw$) are removed.
Masks for discriminative regions were sampled with scales between $0.15$ and $0.2$, ensuring a minimum of 10 patches per mask. 
We set the number of regions $N$=5, the same as I-JEPA. 
Masks are enforced to be non-overlapping.

\paragraph{Alternative Attention Proxy.} We also evaluate a gradient-based attention proxy using a label-free Grad-CAM formulation~\cite{gradcam}. Specifically, we backpropagate the self-supervised prediction objective $\mathcal{L}_{\mathrm{pred}}$ to obtain gradients with respect to the spatial tokens of the last transformer block and derive the attention map following the Grad-CAM procedure:
\begin{equation}
[\mathbf{G}]_{i,j} =
\operatorname{ReLU}\left(
\frac{\partial \mathcal{L}_{\mathrm{pred}}}{\partial [\mathbf{s}_{R_{k+1}}]_{i,j}}
[\mathbf{s}_{R_{k+1}}]_{i,j}
\right).
\label{eq:gradcam}
\end{equation}
This proxy is used only for ablation analysis; our default method uses the \texttt{[CLS]}-based attention.

\paragraph{Computational Resources.}
All experiments were conducted on 8 H200 GPUs with 140 GB of memory. This hardware configuration was used for convenience and is not required for reproducing our results.

\subsubsection{(3) Downstream Evaluation}
\paragraph{Image Classification.} 
We use the ImageNet-1K \cite{imagenet} validation subset for evaluating learned representations in a large-scale benchmark. 
\begin{enumerate}[leftmargin=*,topsep=2pt,itemsep=1pt]
    \item[-] Linear probing: we train a linear head on the full training set, where we concatenate token representations from the last four transformer blocks, following~\cite{ijepa,cjepa}.
    The linear head was configured with batch normalization and optimized using SGD with a momentum of $0.9$, Nesterov momentum, and a weight decay of $5 \times 10^{-4}$. Training was performed for 28 epochs with a base learning rate of $0.01$, scaled automatically with the batch size, and decayed in a multi-step schedule at epochs 8, 16, and 24 ( $0.01 \rightarrow 0.001 \rightarrow 0.0001 \rightarrow 0.00001$). 
    We used a batch size of 32 per replica and applied random resized crop augmentation, resizing images to 256 pixels before cropping to $224 \times 224$.
    \item[-] 
    Fine-tuning: we adapt the fine-tuning protocol of MAE \cite{mae}. Specifically, we fine-tune pre-trained models for 50 epochs on the full training set using the AdamW optimizer and a cosine learning rate scheduler. We use a batch size of 512 and adopt the same learning rate setting as in I-JEPA.
\end{enumerate}

\paragraph{Fine-grained Visual Categorization.}
We evaluate on iNaturalist21 \cite{van2018inaturalist}, CUB-200-2011 \cite{cub}, Stanford-Cars \cite{cars}.
Following the ImageNet linear-probing protocol, we train a linear classifier in same setting. 

\paragraph{Detection and Segmentation.} 
We use MS-COCO 2017~\cite{coco} and ADE20K~\cite{ade20k} for evaluation.
\begin{enumerate}[leftmargin=*,topsep=2pt,itemsep=1pt]
    \item[-]
    Object detection: we fine-tune a Mask R-CNN detector~\cite{maskrcnn} with an FPN on MS-COCO 2017~\cite{coco}, reporting $\mathrm{AP}^{\text{box}}$ and $\mathrm{AP}^{\text{mask}}$ on Val 2017 following the standard COCO protocol. The convolutional backbone is replaced with a ViT-B encoder pretrained using \ours, as in \cite{cjepa,mae}. We fine-tune the entire model end-to-end for 25 epochs with AdamW, a base learning rate of $1.0 \times 10^{-4}$, weight decay $0.1$, using linear warmup in the first epoch followed by cosine decay, a global batch size of $16$. For the ViT-B backbone we use stochastic depth with maximum drop-path rate $0.1$, and otherwise adopt default Mask R-CNN hyperparameters from the underlying detection framework.
    \item[-]
    Semantic segmentation: we adapt a UPerNet-style decoder with a ViT-B backbone pretrained using \ours. We extract token features from a pre-trained model at $3$, $5$, $7$, and $11$ layers, and pass them into decoder. We train and evaluate with an input crop size of $512 \times 512$, and use a sliding-window inference strategy with a $512 \times 512$ crop and a stride of $341 \times 341$. The model is optimized with AdamW with a learning rate of $1 \times 10^{-4}$, a weight decay of $0.05$. 
\end{enumerate}

\paragraph{Low-level Tasks.}
We evaluate \ours~representations on the CLEVR-Count and CLEVR-Distance tasks~\cite{clevr}. 
We use a ViT-based backbone pretrained with \ours~and freeze the target encoder during downstream training. Each image is resized to an input resolution of $224 \times 224$ and fed into the backbone; we take the final-layer class token as a global image representation and pass it through a task-specific linear classification head. CLEVR-Count and CLEVR-Distance are treated as $8$-way and $6$-way classification problems, respectively. 
The linear heads are optimized with AdamW with a learning rate of $1 \times 10^{-3}$, a weight decay of $0.05$, a batch size of $256$, and we train for $100$ epochs using a cosine learning rate schedule.

\begin{figure}[t]
    \centering
    \includegraphics[width=\linewidth]{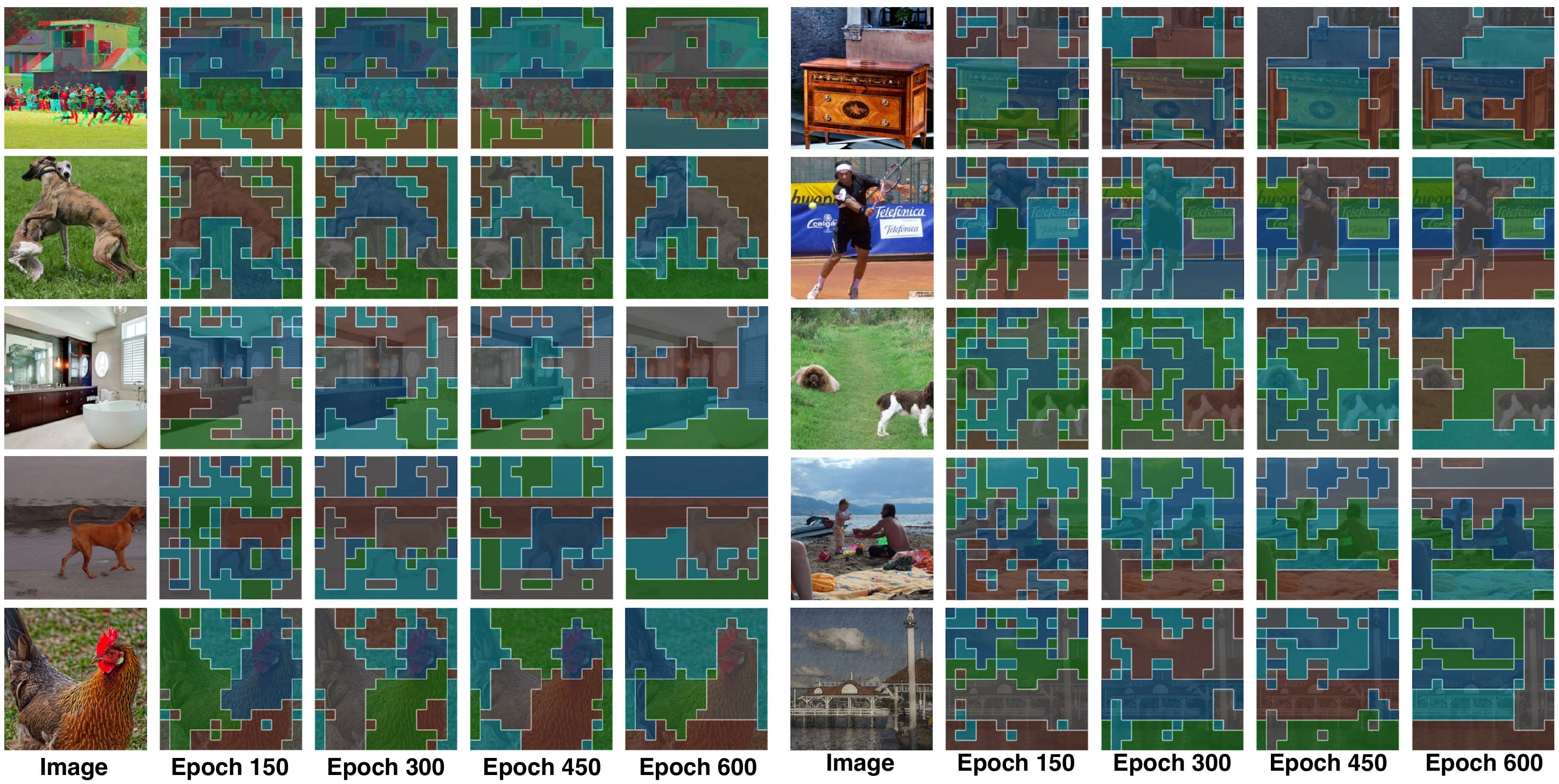}
    \caption{\textbf{Evolution of patch-level clustering during pre-training.} From left to right: input image and 5-cluster results from ViT-B/16 checkpoints at 150, 300, 450, and 600 epochs.}
    \label{fig:patch_cluster_val}
\end{figure}
\subsection{Additional Visualizations.}

To verify that this behavior generalizes beyond the training set and to explore a slightly finer
partition, \Cref{fig:patch_cluster_val} shows analogous visualizations on held-out \emph{test} images with 5 clusters using the ViT-B/16 checkpoints at 150, 300, 450, and 600 epochs. Consistent with the training examples, \ours~encourages patches with similar semantics to be grouped together into compact, coherent clusters. These results further support our claim that discriminative sequencing leads to more structured and interpretable patch-level representations.

\end{document}